%% file: main.tex
\documentclass[conference]{IEEEtran}
\usepackage{times}

\usepackage[numbers]{natbib}
\usepackage{multicol}
\usepackage[bookmarks=true]{hyperref}
\usepackage{amsmath}
\usepackage{amssymb}
\usepackage{algorithm,algorithmic}
\usepackage{url}
\usepackage{color}
\usepackage{graphicx}
\usepackage{balance}

\newcommand{\todo}[1]{}
\renewcommand{\todo}[1]{{\color{red}Todo: {#1}}}
\DeclareMathOperator{\E}{\mathbb{E}}

\DeclareMathOperator*{\argmin}{arg\,min}

\makeatletter
\newcommand\fs@spaceruled{\def\@fs@cfont{\bfseries}\let\@fs@capt\floatc@ruled
  \def\@fs@pre{\vspace{0.5\baselineskip}\hrule height.8pt depth0pt \kern2pt}%
  \def\@fs@post{\kern2pt\hrule\relax}%
  \def\@fs@mid{\kern2pt\hrule\kern2pt}%
  \let\@fs@iftopcapt\iftrue}
\makeatother

\makeatletter
\g@addto@macro\normalsize{%
\setlength\abovedisplayskip{3pt}
\setlength\belowdisplayskip{3pt}
\setlength\abovedisplayshortskip{3pt}
\setlength\belowdisplayshortskip{3pt}
}
\makeatother

\begin{document}

\title{Learning Active Task-Oriented Exploration Policies for Bridging the Sim-to-Real Gap}


\author{\authorblockN{Jacky Liang}
\authorblockA{
Robotics Institute\\
Carnegie Mellon University\\
jackyliang@cmu.edu
}
\and
\authorblockN{Saumya Saxena}
\authorblockA{
Robotics Institute\\
Carnegie Mellon University\\
saumyas@andrew.cmu.edu
}
\and
\authorblockN{Oliver Kroemer}
\authorblockA{
Robotics Institute\\
Carnegie Mellon University\\
okroemer@cmu.edu
}
}


%

\setlength{\textfloatsep}{0.2cm}
\setlength{\floatsep}{0.2cm}

\maketitle
\input{includes/0_abstract.tex}
\IEEEpeerreviewmaketitle

\input{includes/1_intro.tex}
\input{includes/2_rw.tex}
\input{includes/3_method.tex}
\input{includes/4_lqr.tex}

\input{includes/5_experiments.tex}

\input{includes/6_conclusion.tex}
\input{includes/7_ack.tex}

\clearpage
\balance
\bibliographystyle{plainnat}
\bibliography{references}

\clearpage
\input{includes/8_appendix.tex}

\end{document}

%% file: includes/0_abstract.tex
\begin{abstract}
Training robotic policies in simulation suffers from the sim-to-real gap, as simulated dynamics can be different from real-world dynamics.
Past works tackled this problem through domain randomization and online system-identification.
The former is sensitive to the manually-specified training distribution of dynamics parameters and can result in behaviors that are overly conservative.
The latter requires learning policies that concurrently perform the task and generate useful trajectories for system identification.
In this work, we propose and analyze a framework for learning exploration policies that explicitly perform task-oriented exploration actions to identify task-relevant system parameters.
These parameters are then used by model-based trajectory optimization algorithms to perform the task in the real world. 
We instantiate the framework in simulation with the Linear Quadratic Regulator as well as in the real world with pouring and object dragging tasks.
Experiments show that task-oriented exploration helps model-based policies adapt to systems with initially unknown parameters, and it leads to better task performance than task-agnostic exploration.
\end{abstract}

%% file: includes/1_intro.tex
\section{Introduction}

Reinforcement Learning (RL) is a powerful paradigm for training robots to perform complex manipulation tasks in the real world~\cite{kroemer2019review}.
RL methods, whether model-free or model-based, often require a lot of data that is expensive to obtain with real robots.
Instead, many prior works have studied how to train a task policy with simulation data.
However, due to differences in simulation and real-world dynamics as well as observation models, policies trained with simulation data tend to suffer from the \textbf{simulation-to-reality} gap, i.e., the distributional differences between training (simulation) and testing (real-world) data are sufficiently large to degrade the performance of the policy.

Many methods have been proposed to address the sim-to-real gap, including domain adaptation~\cite{bousmalis2018using} and domain randomization for model-free RL~\cite{akkaya2019solving}, and learning residual models that correct sim-to-real errors for model-based RL~\cite{allevato2019tunenet}.
Past works also showed that it is possible to adapt simulation parameters with real-world observations to train model-free RL policies~\cite{chebotar2019closing}, and to use models learned from real-world data to directly perform trajectory optimization for manipulation tasks~\cite{PDDM}.

If a known model with initially unknown parameters is given, System Identification (Sys-Id) can be used to tune these parameters to match the model with different instances of real-world environments.
Sys-Id can be passive or active---the former uses offline trajectories or ones incurred during task execution, and the latter uses an explicit information-gathering exploration policy to probe the environment.
The observation trajectories generated by such an exploration policy are used to fit the unknown parameters of the dynamics model, i.e. the simulator.
Then, a model-based trajectory optimization or planning algorithm uses the model with the estimated parameters to produce a task policy, which is then executed in the real world.

\begin{figure}[!tb]
    \centering
    \includegraphics[width=\linewidth]{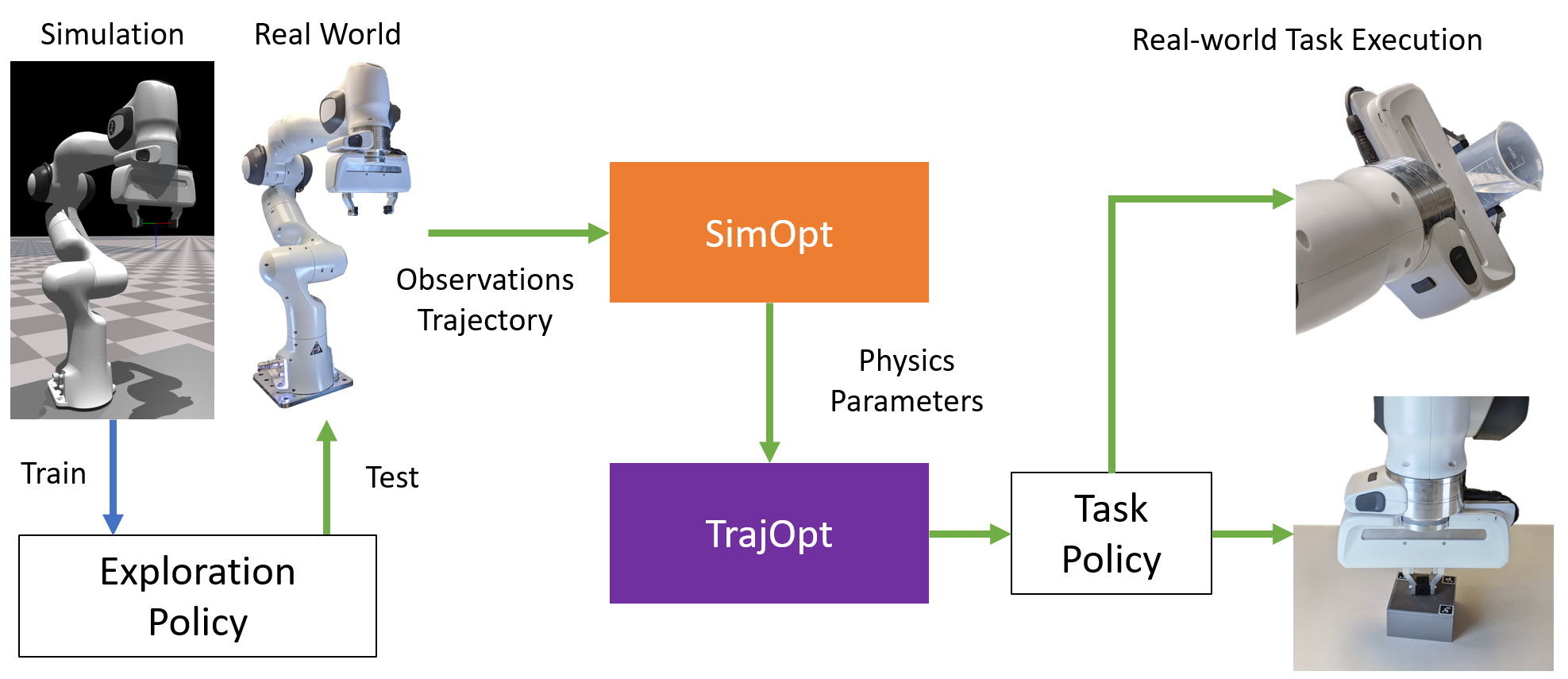}
    \vspace{-15pt}
    \caption{
        Proposed framework of active, task-oriented exploration policies. 
        An exploration policy generates real-world trajectories, which is then used by an optimizer to identify the dynamics parameters (SimOpt). 
        Model-based trajectory optimization uses these parameters to find a task policy (TrajOpt), which then performs the task in the real world. 
        Experiments cover three instantiations of the framework in one simulated task of LQR and two real-world tasks of pouring and object dragging with the Franka Panda robot arms.
    }
    \label{fig:teaser}
\end{figure}

Our proposed approach (Figure~\ref{fig:teaser}) follows this route.
It explicitly learns an exploration policy that interacts with the real world and identifies the initially unknown parameters of a known model, such that task policies planned with that model can succeed in the real world.
For example, the model can be a full dynamics simulation, with the unknown parameters being the mass and friction of the objects involved in the task.

The exploration policies in prior works optimize for model parameter accuracy~\cite{allevato2019tunenet} or model prediction errors~\cite{zhou2019environment}, and as such we call them \textbf{Active Task-Agnostic} exploration.
However, it is usually the case that some parameters are more important to trajectory optimization and task performance than others.
Hence exploration policies that identify all parameters equally well may lead to worse task performance than ones that focus on parameters about which the task is sensitive.

In this work, we propose learning \textbf{Active Task-Oriented} exploration policies, where the exploration policy directly optimizes for the performance of the downstream task.
The exploration policy is trained in simulation over a distribution of physics and task parameters.
Once learned, the exploration policy can be applied in environments with different parameters and different task instances without retraining.
This is in contrast to previous works that adapt simulations to or learn models of a specific instance of real-world dynamics and task parameters, which limit such generalizations.

We performed three experiments to evaluate our framework---one in simulation with the Linear Quadratic Regulator (LQR) and two in real world with pouring and box dragging tasks.
The experiments show that task-oriented exploration helps model-based policies adapt by identifying system parameters and that task-oriented exploration leads to better task performance than task-agnostic exploration.
See videos and supplementary materials at \url{https://sites.google.com/view/task-oriented-exploration/}

%% file: includes/2_rw.tex
\section{Related Works}

One popular method to address the sim-to-real gap is Domain Randomization (DR), which can be applied for observation models~\cite{rusu2016sim, tobin2017domain} or dynamics models~\cite{peng2018sim,muratore2018domain, mehta2019active, mozifian2019learning, ramos2019bayessim}, or both~\cite{akkaya2019solving}.
DR does not aim to train a policy with simulation parameters that is close to those of the real world.
Instead, DR trains the policy with a wide distribution of parameters, with the idea that a policy that can perform the task under all of the different simulations should also be able to perform it in the real world.
As such, DR can make policies more robust.
However, training a policy that works on average for all parameters may lead to sub-optimal performance when different parameters require different policy behaviors~\cite{zhou2019environment, mehta2019active}.
In addition, DR often needs humans to fine-tune the parameter distribution with some prior knowledge about the appropriate range of parameter values in the real world.
The authors of~\cite{mehta2019active, akkaya2019solving} address this issue by effectively building a curriculum that actively changes the DR distributions during training, leading to better policy generalization than sampling from a static and uniform parameter distribution.
The authors of~\cite{chebotar2019closing, ramos2019bayessim} adapt the parameter distribution to better match real-world observations to train model-free RL policies.
However, such adaptation is specific to one instance of real-world environments, so the trained policy is not expected to generalize on environments with different physics or task parameters.

Numerous works have studied using predefined trajectories for Sys-Id~\cite{ogawa2014dynamic}, especially in the context of identifying contact dynamics~\cite{weber2006identification, kolev2015physically, fazeli2017parameter}.
It is also possible to perform manual Sys-Id by directly measuring dynamics parameters in the real world~\cite{tan2018sim}.
However, these methods do not easily scale to different kinds of robots and tasks.
Alternatively, prior works have explored finding the physics parameters, whether for model-free policy learning or model-based adaptations, through trajectories incurred by the task policy.
We refer to these as \textbf{Passive Sys-Id} methods.
For example, the authors of~\cite{yu2017preparing} train a physics parameter-conditioned model-free task policy and a separate prediction model to predict the parameters from a history of trajectories generated by the task policy.
The authors of~\cite{rakelly2019efficient} train an environment-embedding-conditioned RL agent that can identify the embedding online during task execution.
On the model-based side, the authors of~\cite{ross2012agnostic} propose a method to iteratively learn the model in an on-line fashion with trajectories generated by an optimal controller that uses the model, and the algorithm in~\cite{lee2020context} learns a dynamics model conditioned on a local context embedding, extracted by a learned context encoder during task execution.
Passive Sys-Id has the limitation that the trajectory generated by the task policy might not be the most suitable ones for identifying model parameters.

Instead of finding real-world model parameters to improve model accuracy, many works have also studied using real-world data to directly learn the model from scratch or corrections on top of a known but imperfect model.
An example of the former is~\cite{PDDM}, where the authors learn a dynamics model for in-hand manipulation tasks using real-world interactions and perform the task with trajectory optimization.
Works doing the latter are commonly referred to as residual learning~\cite{kloss2017combining, ajay2018augmenting, golemo2018sim, zeng2019tossingbot}.
The method in~\cite{allevato2019tunenet} combines this idea with Passive Sys-Id to produce a model that can iteratively reduce the residual errors with little real-world data.

By contrast to Passive Sys-Id, \textbf{Active Sys-Id} algorithms interact with the environment to explicitly identify relevant system parameters.
This is common for Interactive Perception, where a robot performs probing actions in an environment to segment objects~\cite{van2014probabilistic}, infer object properties like mass~\cite{kannabiran2019estimating}, or infer kinematic constraints~\cite{hausman2015active, baum2017opening, eppner2018physics}.
Like our approach, many of these works assume a known model with unknown parameters (e.g. the type of joint connecting two rigid bodies), and they aim to choose the most informative actions via heuristics like maximum information gain to reduce dynamics prediction error.
The work in~\cite{zhou2019environment} learns an exploration policy that generates informative trajectories for inferring environment embeddings.
These embeddings are used as input to model-free RL policies, and the embedding encoder is trained to reduce dynamics prediction error.
In~\cite{niranjan2019estimating} the authors train an exploration policy that optimizes for the accuracy of a parameter prediction network that estimates the mass of articulated objects.
We describe Active Sys-Id methods like these as Task-Agnostic, because the exploration is done to optimize for the accuracy of all of the model parameters or model predictions, and not a downstream task.

Our work considers the case for performing Active and Task-Oriented exploration, where the exploration policy is used to infer model parameters in a way that directly optimizes for task performance.
The algorithm proposed in~\cite{llamosi2014experimental} is similar---it applies a Bandits-based approach to select from a fixed set of experiments to perform in the environment.
Like~\cite{zhou2019environment} however, this work focuses on the case of a finite set of environments, so the exploration policy only needs to produce information that can \textit{classify} which environment the agent is in, and not the underlying system parameters.
The authors of~\cite{farahmand2018iterative} propose learning a model from scratch that minimizes prediction error in the value function, thus making model-learning task-oriented.
In this work, we assume a known parameterized model and explicitly try to identify the continuous task-relevant system parameters.
This means that the method does not need a predefined discrete set of environments.
It also means the method does not require policy learning or model learning from scratch, because the tuned model is directly used for trajectory optimization to perform the downstream task.

%% file: includes/3_method.tex
\section{Method Overview}
\begin{figure*}[!t]
    \centering
    \includegraphics[width=0.95\linewidth]{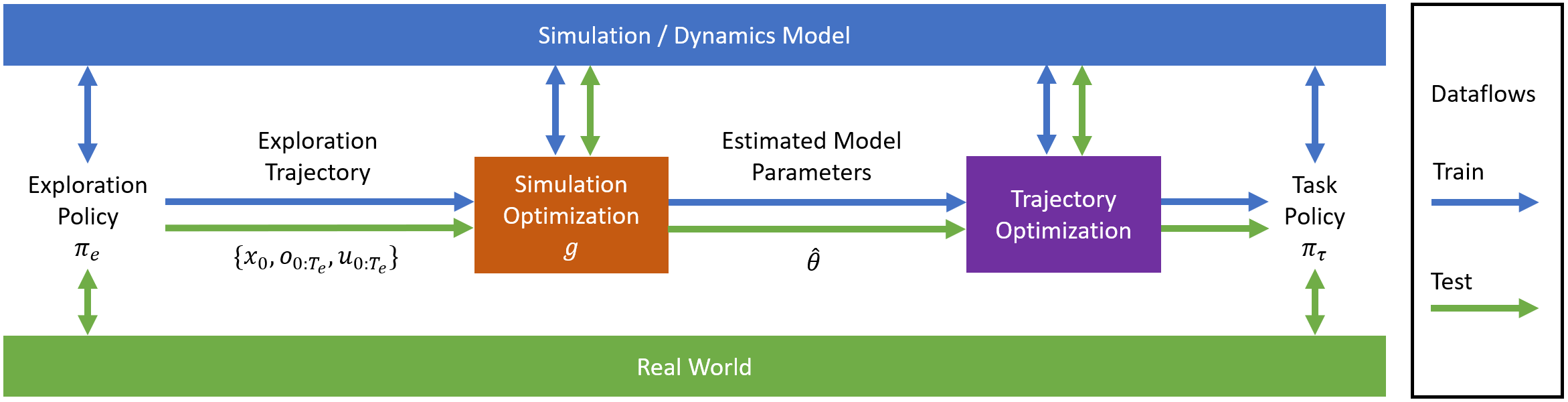}
    \vspace{-5pt}
    \caption{
        Learning active task-oriented exploration policies: training and testing pipeline.
        During training, the $\pi_e\rightarrow\pi_\tau$ pipeline is executed multiple times in simulation to evaluate the expected task regret $\E_{\theta} \Psi$. 
        This is then used to form the objective in Equation~\ref{eq:pi_e_obj} for optimizing $\pi_e$.
        During testing (deploying the learned exploration policy), $\pi_e$ and $\pi_\tau$ interfaces with the real world instead of the simulated dynamics.
    }
    \vspace{-10pt}
    \label{fig:framework}
\end{figure*}

We first give a general overview of the proposed framework, seen in Figure~\ref{fig:framework}, for learning active and task-oriented exploration policies.
Following this section are three instantiations of the framework---one simulated Linear Quadratic Regulator (LQR) task and two real-world manipulation tasks.

\subsection{Active and Task-Oriented Exploration}

We consider the problem of fitting the parameters of a dynamics model by using an exploration policy to generate interactions with the real system, so that a  task policy planned with the fitted dynamics model can be directly applied on the real system with no further fine-tuning.

Our chief assumption is that for a given task there exists some values of model parameters that will make the simulated dynamics sufficiently match real-world dynamics, so a task policy planned using the estimated parameters can also complete the task in the real world. 
We further assume that it is possible to recover these parameter values from observations available to the robot, and that we have models of the objects, the robot, as well as their initial states.

We define the following variables:
\begin{itemize}
    \item $x_t$ - state (e.g. robot, object configurations).
    \item $u_t$ - robot actions (e.g. desired end-effector poses)
    \item $o_t$ - observations (e.g. tracked object trajectories)
    \item $\theta$ - dynamics model parameters that can affect task execution (e.g. mass, friction).
    \item $f_\theta(x_t, u_t)$ - a discrete-time dynamics function that yields $x_{t+1}$ using the parameters $\theta$. This is used to describe both simulated and real-world dynamics functions. The simulated dynamics is deterministic, while real-world dynamics is stochastic due to unmodeled noise in the environment.
\end{itemize}

\textbf{Model-based Task Policy.}
Let $\pi_\tau(\theta)$ be a task policy that uses the dynamics model with model parameters $\theta$.
For example, it could be a policy that performs a motion plan generated by trajectory optimization with a physics simulator as the dynamics model.
Let $T_\tau$ be the time horizon of the task.
Forming $\pi_\tau$ is called Trajectory Optimization (TrajOpt).

Let $J(\pi_\tau(\hat{\theta}), \theta)$ be the total cost of a task performed by a policy based on $\hat{\theta}$ acting in an environment that actually has physics parameters $\theta$.
$J$ can be an expectation over a distribution of tasks (e.g. a distribution over goal states), but for brevity we omit the expectation symbol $\E$ when writing $J$.
The expected loss over $\theta$ is $\E_\theta J(\pi_\tau(\hat{\theta}), \theta)$.
If the policy is optimized for one set of parameters but deployed on a system that has different parameters, we expect the performance of the policy to be worse than one with the parameters it was optimized for: $\E_{\theta}J(\pi_\tau(\hat{\theta}), \theta) \ge \E_{\hat{\theta}}J(\pi_\tau(\hat{\theta}), \hat{\theta})$.

\textbf{Simulation Parameter Optimization.}
An exploration trajectory consists of an initial state $x_0$, real-world observations $o_{0:T_e}$, and actions $u_{0:T_e}$, where $T_e$ is the horizon for the 
exploration trajectory.
Let $g$ be the optimizer that optimizes for the physics parameters that match these trajectories: $g(x_0, o_{0:T_e}, u_{0:T_e}) = \hat{\theta}$.
We call applying $g$ Simulation Optimization (SimOpt).
For example, $g$ can be a closed-form expression that directly solves for $\theta$ or a derivative-free optimization algorithm that iteratively searches for it.
In general, $g$ tries to give an estimate $\hat{\theta}$ that minimizes the prediction error of the resultant dynamics model on the exploration trajectory.

\textbf{Active Task-oriented Exploration Policy.}
Let $\pi_e$ be the task-oriented exploration policy, which acts as a feedback controller starting from an initial state that is given or learned.
The exploration trajectory $\{u_{0:T_e}, o_{0:T_e}\}$ generated by $\pi_e$ is given to $g$, yielding estimated model parameters $\hat{\theta}$ that are used by the task policy $\pi_\tau(\hat{\theta})$ to perform the task.
We call this the $\pi_e\rightarrow\pi_\tau$ pipeline.

The goal of $\pi_e$ is to generate an exploration trajectory that leads to low expected costs incurred by $\pi_\tau$:
\begin{equation} \label{eq:pi_e_obj}
    \pi_e = \argmin_{\pi_e} \E_{\theta}[\Psi(\pi_\tau(\hat{\theta}), \theta) + \gamma h(\pi_e)]
\end{equation}
where $h(\pi_e)$ is a regularization term that penalizes $\pi_e$ from being too complex and incurring states and actions that have high costs, and $\Psi$ denotes the regret of $\pi_\tau(\theta^*)$ w.r.t. a task policy optimized using the real dynamics parameters:
\begin{equation}
    \Psi(\pi_\tau(\hat{\theta}), \theta) = J(\pi_\tau(\hat{\theta}), \theta) - J(\pi_\tau(\theta), \theta)
\end{equation}


See Algorithm~\ref{alg:deploy} for the function that deploys the $\pi_e\rightarrow\pi_\tau$ pipeline at ``test time" to perform a task.
The deploy function takes in a dynamics model $f_{\theta}$, which is the real-world dynamics when the robot is actually performing the task and the simulated dynamics during training.

\floatstyle{spaceruled}
\restylefloat{algorithm}
\begin{algorithm}[!tb]
\caption{Deploy Exploration and Task Policy}
\label{alg:deploy}
\begin{algorithmic}[1]
    \renewcommand{\algorithmicrequire}{\textbf{Input:}}
    \renewcommand{\algorithmicensure}{\textbf{Output:}}
    \REQUIRE $\pi_e$, $f_{\theta}$
    \STATE Roll out $\pi_e$ in environment with dynamics $f_{\theta}$
    \STATE Obtain exploration trajectory $[x_0, o_{0:T_e}, u_{0:T_e}]$ 
    \STATE Estimate model parameters $\hat{\theta} \leftarrow g(x_0, o_{0:T_e}, u_{0:T_e})$
    \STATE Form $\pi_\tau(\hat{\theta})$ via trajectory optimization using $f_{\hat{\theta}}$
    \STATE Roll out $\pi_\tau$ in $f_{\theta}$
    \RETURN $J(\pi_\tau(\hat{\theta}), \theta)$
\end{algorithmic}
\end{algorithm}

This procedure is active, because it chooses how to interact with the real system to generate informative observations for estimating model parameters.
It is also task-oriented as opposed to task-agnostic, because $\pi_e$'s goal is to minimize task regret $\Psi$, and not the accuracy of $g$'s predictions.

\subsection{Training a Task-Oriented Exploration Policy}
The exploration policy $\pi_e$ is trained in simulation via RL, where the reward function is the negative of the objective in Equation~\ref{eq:pi_e_obj}.
At each training iteration we sample ``ground-truth" physics parameters from a wide, predefined distribution $\theta \sim \Theta$.
Then we go through the $\pi_e\rightarrow\pi_\tau$ pipeline, deploy the task policy $\pi_\tau$ in the ``ground-truth" simulation $f_{\theta}$, and evaluate the task regret $\Psi$. 
If there is a distribution of task objectives (e.g. goal object poses), we repeat this process with multiple task samples.
This constitutes one rollout of $\pi_e$.
The reward for $\pi_e$ is sparse, as it can only be computed after the entire exploration trajectory is performed.
See Algorithm~\ref{alg:eval} for how to evaluate the objective's expected regret term, $\E_{\theta} \Psi(\pi_\tau(\hat{\theta}), \theta)$, in simulation during training.

Sampling $\theta$ from a distribution, instead of training $\pi_e$ for a specific $\theta$, is motivated by works in DR.
We apply DR's argument not to the task policy, but to the exploration policy; $\pi_e$ should be applicable to a wide range of simulated environments, and if it is, then it should also be able to work in the real world even if it was trained in simulation.
This should be easier to achieve than with the task policy, because the space of trajectories that completes a task is more restrictive than one that's informative for Sys-Id. 

Optimizing for the expected task regret $\E_{\theta} \Psi(\pi_\tau(\hat{\theta}), \theta)$ w.r.t. $\pi_e$ is equivalent to optimizing for the expected task performance $\E_{\theta} J(\pi_\tau(\hat{\theta}), \theta)$.
This is because the the difference, the expected task performance of the policy using ground-truth system parameters $\E_{\theta} J(\pi_\tau(\theta), \theta)$, does not depend on the exploration policy.
In practice, we optimize for $\E_{\theta} J(\pi_\tau(\hat{\theta}), \theta)$, but during training we report $\E_{\theta} \Psi(\pi_\tau(\hat{\theta}), \theta)$, because $\Psi$ fluctuates less than $J$ with noisy dynamics, and $\Psi = 0$ is an interpretable target for training $\pi_e$.

\floatstyle{spaceruled}
\restylefloat{algorithm}
\begin{algorithm}[!tb]
\caption{Evaluate Expected Regret of Exploration Policy}
\label{alg:eval}
\begin{algorithmic}[1]
    \renewcommand{\algorithmicrequire}{\textbf{Input:}}
    \renewcommand{\algorithmicensure}{\textbf{Output:}}
    \REQUIRE $\pi_e$, $\Theta$, $N$
    \FOR{$n\in\{1\hdots N\}$}
        \STATE Sample $\theta_n\sim\Theta$
        \STATE Form $\pi_\tau(\theta_n)$ via trajectory optimization using $f_{\theta}$
        \STATE Evaluate $J(\pi_\tau(\theta_n), \theta_n)$
        \STATE $J(\pi_\tau(\hat{\theta}_n), \theta_n) \leftarrow $ Deploy($\pi_e$, $f_{\theta_n})$
        \STATE $\Psi_n \leftarrow J(\pi_\tau(\hat{\theta}_n), \theta_n) - J(\pi_\tau(\theta_n), \theta_n)$ 
    \ENDFOR
    \RETURN $\E_{\theta} \Psi \approx \frac{1}{N}\sum_{n=1}^N\Psi_n$
\end{algorithmic}
\end{algorithm}

%% file: includes/4_lqr.tex
\section{Analysis and Experiments for LQR}

To better understand the proposed framework, we first apply it to the LQR task, which is fast to train and amenable to analysis.
The task is to form a linear feedback controller to act in a fully observable discrete-time linear system to minimize a finite-horizon quadratic cost in terms of the states and actions:
\begin{align}
    x_{t+1} &= Ax_t + Bu_t \\
    J &= \sum_{t=1}^{T_\tau}{x_t^{\top}Qx_t + u_t^{\top}Ru_t}
\end{align}
where $x_t \in \mathbb{R}^n$, $u_t \in\mathbb{R}^m$, $Q\in\mathbb{R}^{n\times n}$, and $R\in\mathbb{R}^{m\times m}$.
$Q$ and $R$ are positive semi-definite cost matrices.

We set $A$ to have the form $A = U \theta U^\top$, where $U$'s columns are the eigenvectors of $A$ and $\theta$ the corresponding diagonal matrix of eigenvalues.
This allows us to decompose the system dynamics into parts that are already known ($U$) and the parts that are unknown ($\theta$).
The parameters that the exploration policy tries to infer are $\theta$, the eigenvalues of $A$.

\subsection{LQR Analysis}

Our goal in this section is to rewrite the minimization of the task cost $\min J$ as an optimization problem w.r.t to an exploration policy $\pi_e$ and see how the task-oriented objective affects the optimization.
In the equations that follow, we denote $A$ as the ground-truth system dynamics, and $\hat{A}$ as the estimated system dynamics.
The cost $J$ is computed with respect to $A$, while the model-based LQR policy $\pi_\tau$ is computed with respect to $\hat{A}$.
We assume $B$ is known.

\textbf{Trajectory Optimization.}
With the task of minimizing $J$, the optimal LQR policy is a linear feedback controller with time-varying gains $u_t = K_{\tau, t} x_t$, where the gains are computed as follows:

\begin{align}
    P_{T_\tau} &= Q \\
    K_{\tau, t} &= -(R + B^\top P_{t+1}B)^{-1}B^\top P_{t+1}\hat{A} \\
    P_t &= Q + K_{\tau, t}^\top R K_{\tau, t} + (\hat{A} + BK_{\tau, t})^\top P_{t+1}(\hat{A} + BK_{\tau, t})
\end{align}


\textbf{Simulation Optimization.}
With linear dynamics and full observability, we can derive the closed form solution for $\hat{\theta}$ given an exploration trajectory by using the following objective:
\begin{align}
    \hat{\theta} &= g(x_{0:T_e}, u_{0:T_e}) = \argmin_\theta \sum_{t=1}^{T_e} \|\hat{x_t} - x_t\|_2^2
\end{align}
Taking the derivative and setting it equal to $0$ yields:
\begin{align}
    \hat{\theta} &= \argmin_\theta \sum_{t=1}^{T_e} \|U\theta U^\top x_{t-1} + Bu_{t-1} - x_t\|_2^2\\
    &= -U\big(\sum_{t=1}^{T_e} x_t x_t^\top\big)^\dagger U^\top \big(\sum_{t=1}^{T_e}(U^\top x_t)\circ(U^\top(Bu_{t-1} - x_t))\big)
\end{align}
where $\circ$ denotes element-wise product and $\dagger$ the pseudo-inverse.
If $T_e \ge n$ and $[x_1 \hdots x_{T_e}]$ spans $\mathbb{R}^n$, then $\sum_{t=1}^{T_e} x_t x_t^\top$ is invertible.
In practice, we use the pseudoinverse to handle edge cases and also resolve numerical instabilities.



\textbf{Task-Oriented Exploration Policy.}
We set the exploration policy to be a linear feedback controller with time-invariant gains: $u_t = K_e x_t$.
Under the proposed framework, $\pi_e$ will generate a trajectory of length $T_e$, which will be used to estimate $\hat{A} = U\hat{\theta}U^\top$.
The estimated dynamics model is used to compute the optimal LQR gains, the performance of which will be evaluated in $J$.

To make the analysis tractable, we make the following assumptions: $n=m$, $B=I$, $Q$ is diagonal, and $R=0$.
Setting $R=0$ is the strongest assumption, and it is not met in practice, as doing so gives zero penalities to arbitrarily large controller effort (although the closed loop system does not converge in $1$ step if there are modeling errors $\|\hat{A} - A\| > 0$).
However, having $R=0$ greatly simplifies the algebra that follows, and the result still provides useful insights.

With these assumptions the LQR equations yield $P_{T_\tau} = Q, K_t = -\hat{A}, P_T = Q$.
We can rewrite the costs in terms of the closed-loop dynamics by using this simplified LQR policy:
\begin{align}
    x_{t+1} &= (A + B K_{t}) x_{t} = (A - \hat{A}) x_{t}\\
    J &= \sum_{t=1}^{T_\tau} x_t^{\top}Qx_t = \sum_{t=1}^{T_\tau} x_{t-1}^\top(A - \hat{A})^\top Q(A-\hat{A})x_{t-1}
\end{align}
Finding the optimal $K_e$ to minimize $J$ can be formulated as an optimization problem on setting the gradient $\nabla_{K_e} J$ to $0$:
\begin{align}
    K_e^* &= \argmin_{K_e} \|\nabla_{K_e}J\| \\
    &= \argmin_{K_e} \|\nabla_{K_e}\hat{\theta} \nabla_{\hat{\theta}} J\|  \\
    &= \argmin_{K_e} \|\nabla_{K_e}\hat{\theta} \sum_{t=1}^{T_\tau} U^\top(D_t + D_t^\top)U\| \label{eq:lqr_obj}
\end{align}
where $D_t = x_{t-1}x_{t-1}^\top(\hat{A} - A)Q = x_{t-1}(\hat{x}_t - x_t)^\top Q$.

Note that the objective function of optimizing the task cost $J$ w.r.t. the exploration policy $K_e$ is a combination of 
1) how sensitive the identified parameters $\hat{\theta}$ are to the exploration policy ($\nabla_{K_e}\hat{\theta}$), 
2) the dynamics prediction error (the $\hat{x}_t - x_t$ term of $D_t$) weighted by
3) the task costs (the $Q$ term of $D_t$).
As $K_e$ depends on all $3$ of these factors, this analysis on the simplified system illustrates the difference between task-oriented system identification vs. the task-agnostic variant, which would not have terms that depend on task performance.

\subsection{LQR Simulation Experiment}
\begin{figure}[!tb]
    \centering
    \includegraphics[width=0.9\linewidth]{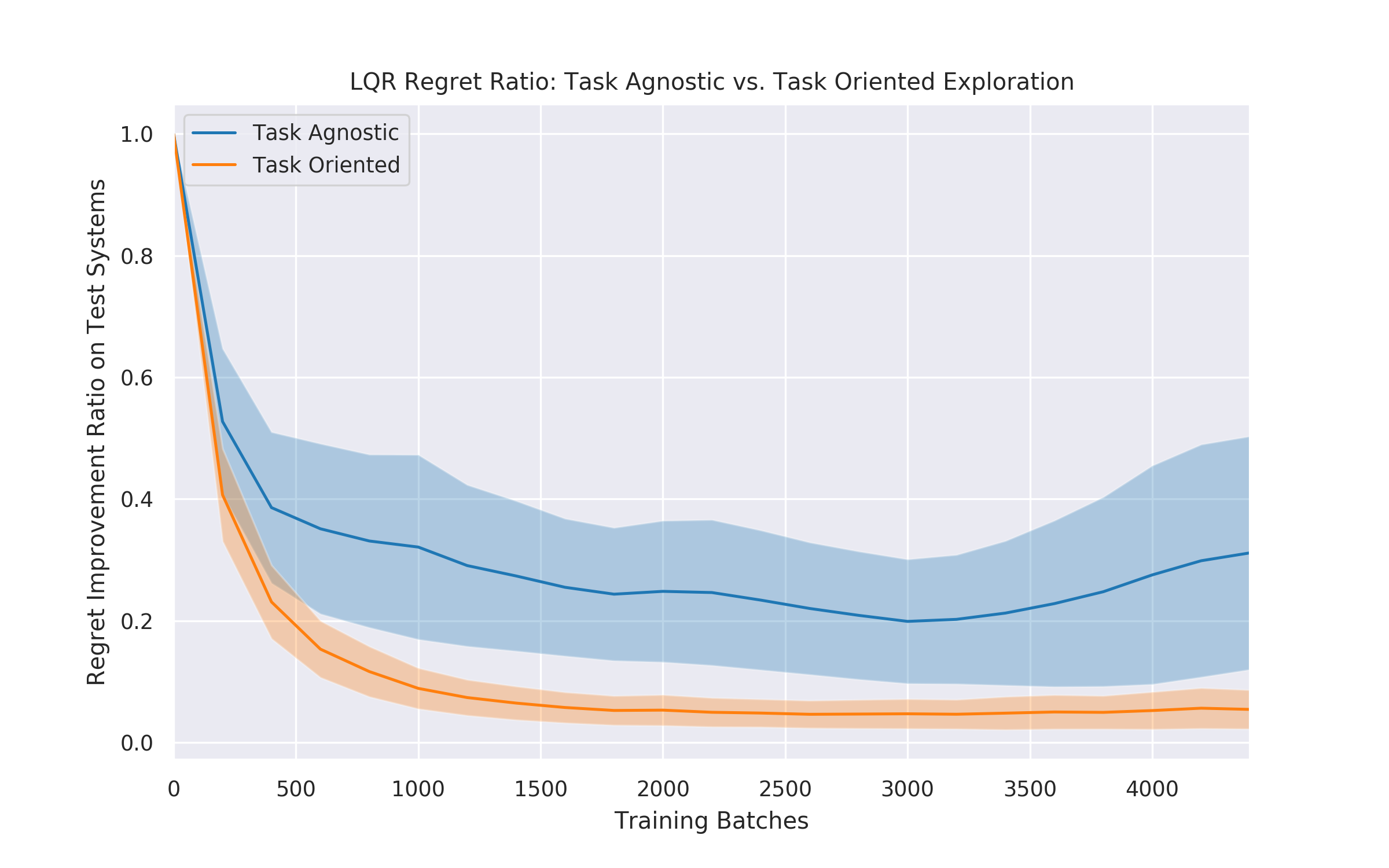}
    \vspace{-15pt}
    \caption{
        Comparison of task-agnostic vs. task-oriented exploration policy training in regret ratio of LQR cost on test systems across $10$ random seeds.
        Task-oriented exploration achieves lower final regret ratio, converges faster, and incurs lower variance on task regret than task-agnostic exploration.
    }
    \label{fig:lqr_regret_ratio}
\end{figure}

We implemented the proposed framework with a linear system and the LQR task as the previous section describes.
Notably, our experiments do not make the simplifying assumptions that the analysis makes, with the exception of the form of the system $A = U\theta U^\top$ and that $U$ and $B$ are known.

We used gradient descent to optimize $J$ w.r.t. $K_e$.
Evaluating the LQR costs, obtaining the optimal discrete-time LQR policy, and obtaining the estimate $\hat{A}$ are all  differentiable (the last two are differentiable by using their closed form solutions).
As such, the entire pipeline from the exploration policy to evaluating LQR costs is end-to-end differentiable.

To sample $A$, we first randomly generate a fixed orthonormal basis $U$, then we sample eigenvalues $\theta \sim \mathcal{N}(\mu, \sigma I)$. 
In our experiment, we used $n=6$, $m=3$.
The sampled eigenvalues are capped at a magnitude of $1.1$, so the systems have slightly unstable open-loop behavior, which makes LQR non-trivial.
The system also has small amounts of observation and dynamics noise, both sampled from i.i.d. zero-mean isotropic normal distributions at every time step.

The training set contains $1000$ examples of $\theta$, with the test set containing $100$.
Gradient descent was done by the Adam optimizer with a learning rate of $10^{-4}$ and weight decay of $0.1$.
We also put an LQR-like cost on the trajectory generated by the exploration policy: $h(\pi_e) = \sum_{t=1}^{T_e} x_t^\top Q_e x_t + u_t^\top R_e u_t$.
The task policy horizon is $20$, while the exploration policy horizon is $4$.
Initial state $x_0$ for the task is fixed, while initial state for exploration is optimized for along with the exploration policy's feedback gains.

We compare the proposed task-oriented exploration policy vs. a baseline task-agnostic exploration policy.
The task-oriented policy is trained to minimize regret of the task policy $E_\theta \Psi$, while the task-agnostic exploration policy is trained to minimize parameter estimation error $E_\theta \|\hat{\theta} - \theta\|_2^2$.

Note that the task-agnostic exploration policy is not optimizing for the model prediction accuracy on the exploration trajectory.
Doing so would lead the exploration policy to generate trajectories that are easy to predict.
In some systems, this may lead to the policy doing nothing, incurring no state changes, and hence predictions become trivial.
The fitted dynamics parameters in this case would not be useful for downstream tasks.

Figure~\ref{fig:lqr_regret_ratio} plots the results of this experiment.
The x-axis denotes the number of training batches.
The y-axis denotes the ratio between the regret achieved by the exploration policy on test systems vs. the regret of the initial random exploration policy, which is the same for both task-agnostic and task-aware training runs.
Regret ratio is reported here, because the unitless LQR cost is difficult to interpret, and we can compute the optimal regret and provide a more intuitive value between 0 and 1.
We ran the training procedure for $10$ random seeds, and the means and standard deviations are computed across those seeds.

While both task-agnostic and task-oriented exploration policies are able to reduce regret, the task-oriented exploration performs better than the task-agnostic variant by having faster convergence toward a lower final regret ratio, as well as having a smaller variance.

%% file: includes/5_experiments.tex
\section{Real-world Robot Experiments}

We apply the framework to two real-world robot manipulation tasks, one using an analytical model and a discrete exploration action space, and one using full dynamics simulations with continuous exploration action space.

\subsection{Task: Pouring}

In the pouring task, the robot must pour $m_\tau$g of water from a cup with known shape but containing initially unknown amount of water.
The goal parameter $m_\tau$ is sampled at every execution of the task: $m_\tau\sim \mathcal{N}(\mu_{m_\tau}, \sigma_{m_\tau})$.
Because the cup shapes are known a priori, if we know the initial amount of water in each cup, we can compute the exact angle at which to tilt the cup to pour the desired amount $m_\tau$.
The task policy $\pi_\tau$ in this case has just one parameter---the cup tilt angle $\phi$ ($\phi = 0$ when the cup stands upright and $\phi = \pi/2$ when the cup is laying horizontal).
Let $\hat{m}_\tau$ refer to the actual amount of water poured. 
The task cost is $J = |m_\tau - \hat{m}_\tau|$.

Below is the analytical solution that relates the tilt angle $\phi$ of a cylindrical cup with uniform radius $r$, the height $h$, and the maximum volume $V$ of fluid that remains in the cup:
\begin{equation}
    \phi = \tan^{-1}(\frac{1}{r}(h - \frac{V}{\pi r^2}))
\end{equation}
This equation is used to compute $\phi$ given a desired $V$, calculated from the amount of water that should be left in cup after pouring ($m_c - m_\tau$).
Note the above model only works when $\phi < \tan^{-1}(\frac{h}{2r})$, and we enforce this constraint during experiments.

In addition to the cup the task uses, another identical distractor cup is also in the scene.
The unknown system parameter are the initial masses of both of the cups $\theta = [m_1, m_2]$.
The exploration policy $\pi_e$ operates in a discrete action space---lifting either cup $0$ or cup $1$ and uses the end-effector force measurements to estimate the mass of the lifted cup.
This measurement is noisy, so lifting a cup more times result in a more accurate mass measurement, which would in turn lead to smaller task costs.
The exploration policy acts as follows---it first performs a single measurement for both of the cups.
For the remaining $T_e - 2$ time steps, it samples which cup to lift from a Bernoulli distribution with parameter $p_e$, where a value of $1$ means choosing the task-relevant cup, and $0$ the task-irrelevant cup.
Ideally, a trained exploration policy has a $p_e$ that strongly favors the cup the task uses, leading to a more accurate initial mass estimate and better task performance.

We trained $\pi_e$ by sampling from the analytical model with added observation noise to the mass measurements as well as dynamics noise to the outcome of how much water was poured for a given tilt angle.
$\pi_e$ is optimized via gradient descent with finite-difference approximations of $\frac{\partial \E_\theta J}{\partial p_e}$.
We set $T_e = 6$, so the maximum number of measurements per cup is $5$.

\begin{figure}[!t]
    \centering
    \includegraphics[width=0.49\linewidth]{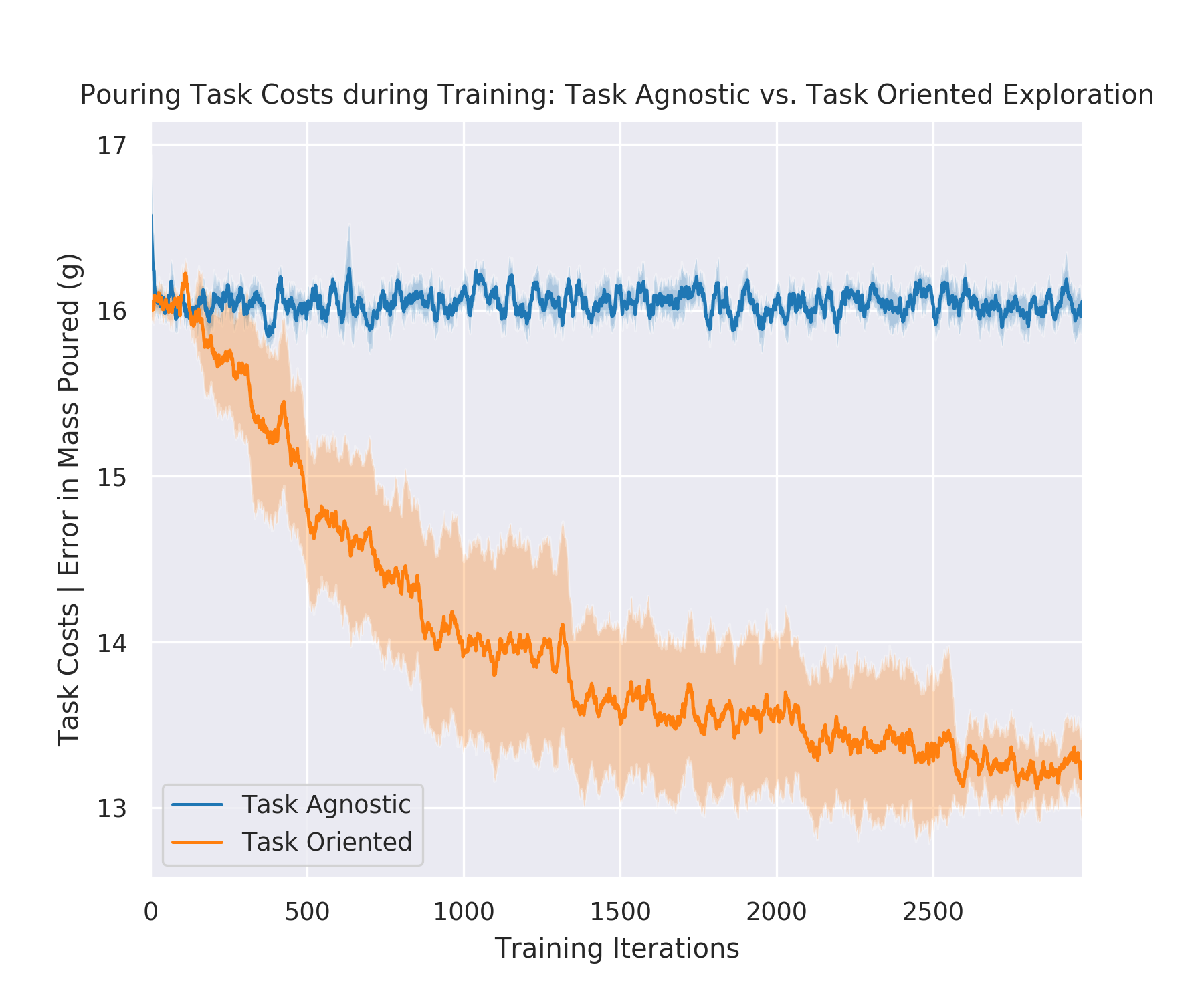}
    \includegraphics[width=0.49\linewidth]{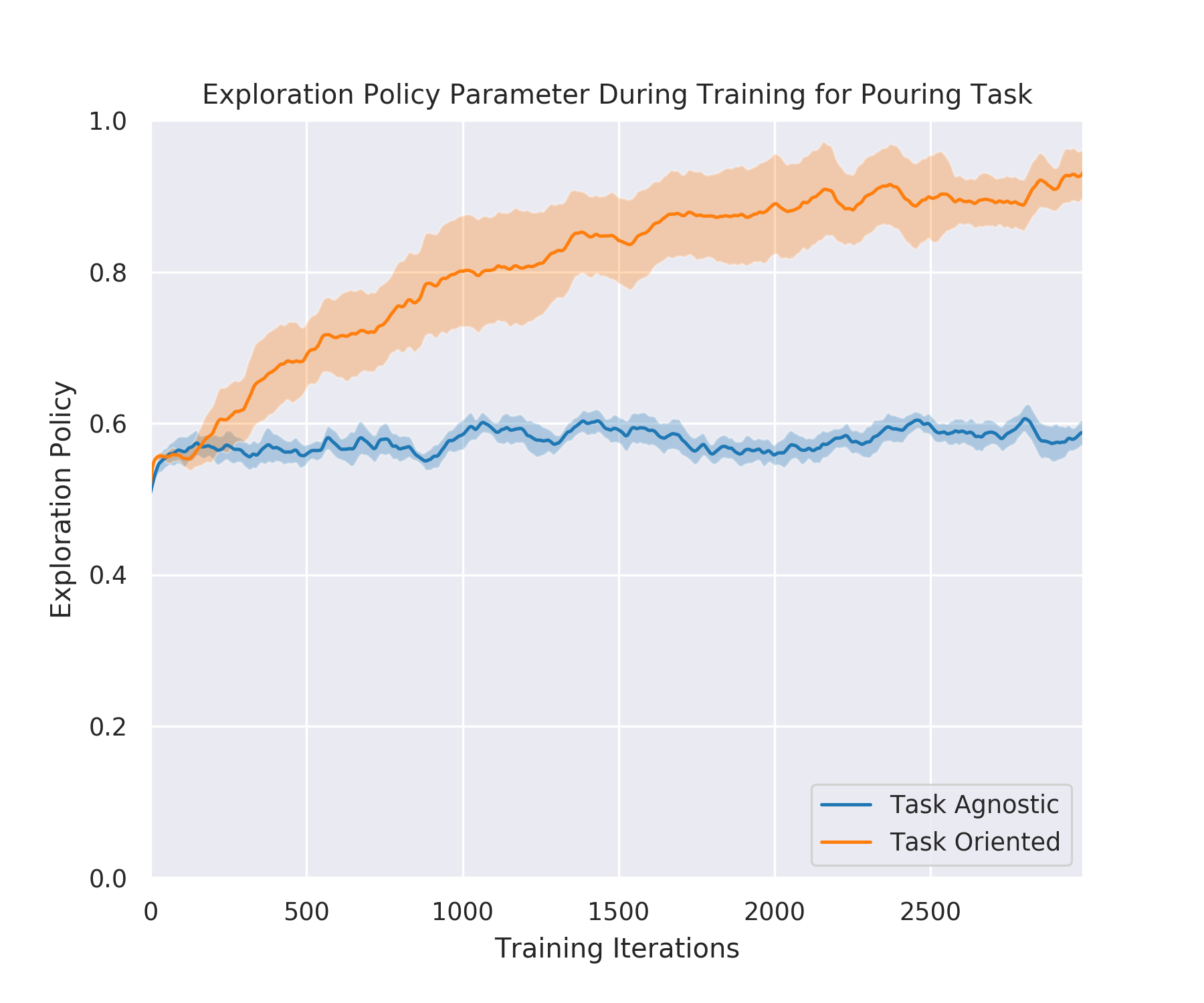}
    \vspace{-15pt}
    \caption{
        Comparison of Task-Agnostic and Task-Oriented exploration policies during training for the pouring task across $10$ random seeds.
        Left: Costs during training. 
        Right: Policy parameter (probability of measuring the cup used by the task).
        Because the Task-Oriented policy is optimizing for task performance and not parameter prediction error, it favors weighing the cup used by the task instead of both cups equally.
    }
    \label{fig:pouring_training}
    \vspace{-5pt}
\end{figure}

Figure~\ref{fig:pouring_training} shows both the costs incurred by the task-oriented and task-agnostic exploration policies and their $p_e$'s during training.
The task-agnostic $p_e$ is around $0.6$, while the task-oriented $p_e$ is closer to $0.9$.
The task-oriented exploration policy also achieves better final task performance.

\begin{figure}[!t]
    \centering
    \includegraphics[width=0.38\linewidth]{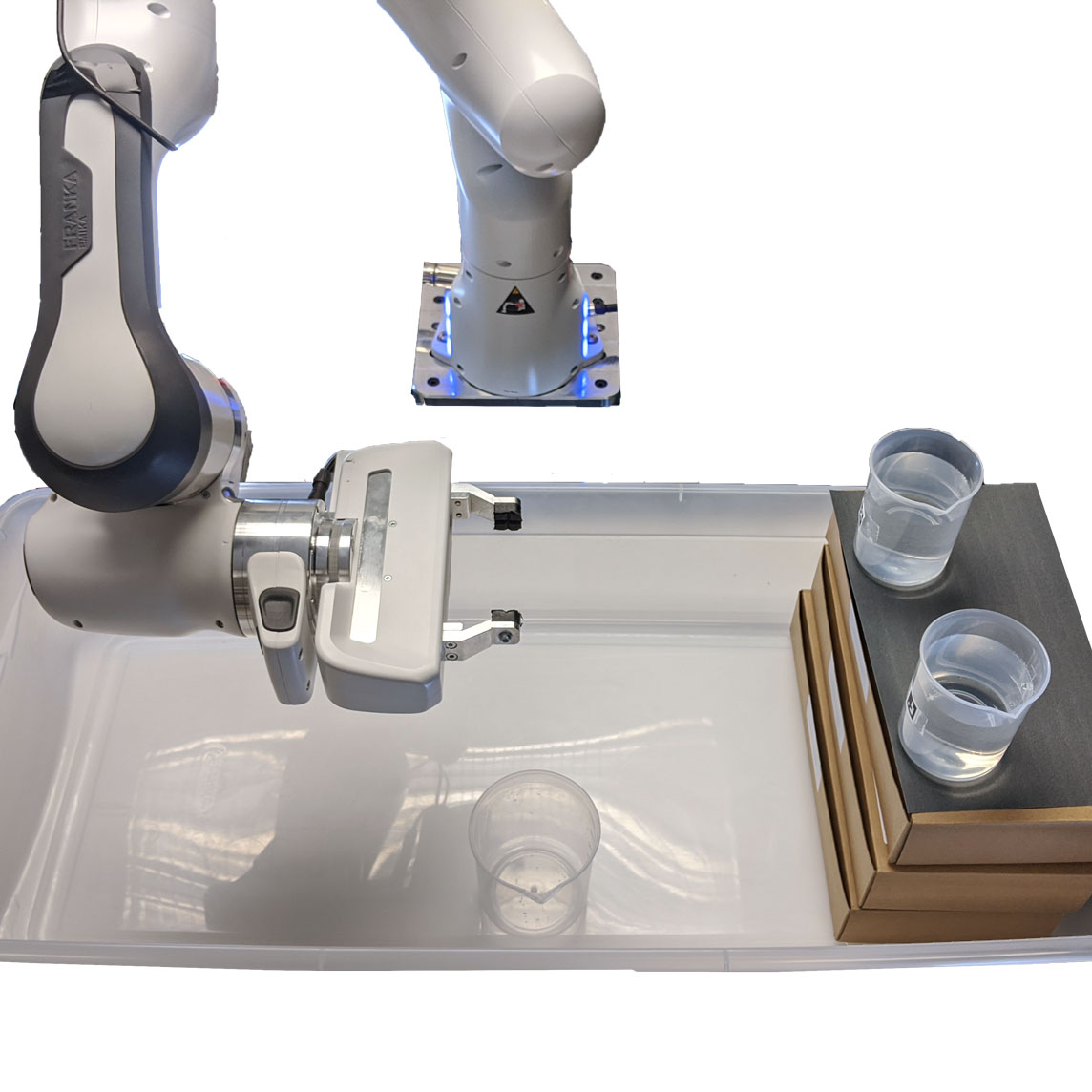}
    \includegraphics[width=0.6\linewidth]{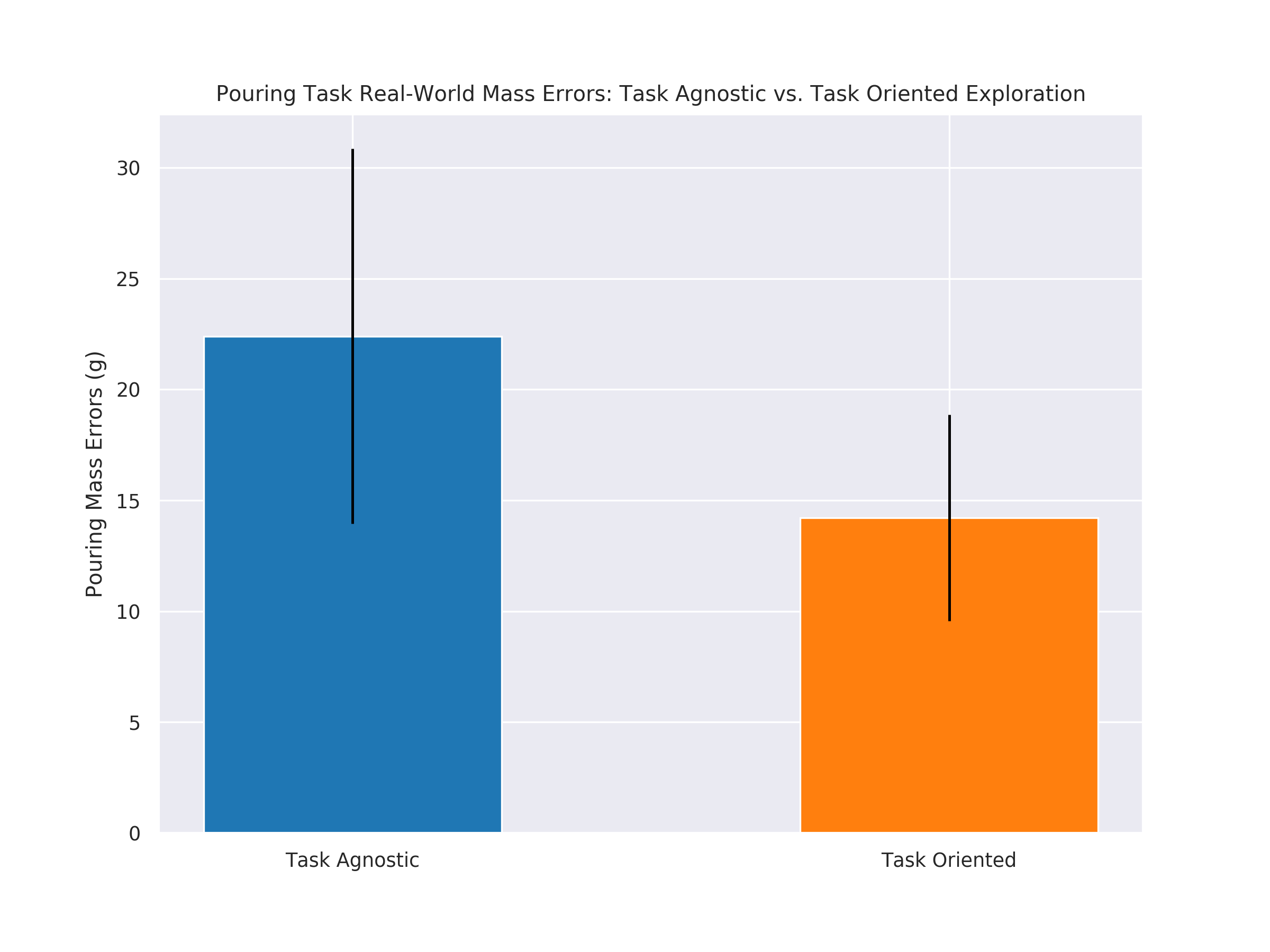}
    \vspace{-25pt}
    \caption{
        Real-world pouring experiment.
        Left: Experiment setup. The robot first to measures the initial masses of the two cups, then it pours a target amount of water from the fixed task-relevant cup into another container.
        Right: Comparison of Task-Agnostic and Task-Oriented exploration policies for real-world pouring task costs. 
    }
    \label{fig:pouring_real_world}
\end{figure}

We performed the pouring task in the real world with a 7 DoF Franka Panda arm.
Two identical plastic beakers were used for the cups, and the masses before and after pouring were measured with a scale.
We evaluated the trained task-oriented and task-agnostic policies with $10$ samples of goal parameters in the real world.
See Figure~\ref{fig:pouring_real_world} for the real-world experiment setup and results.
The task-oriented exploration policy, by focusing exploration on the cup more likely to be used by the task, achieves a lower average task cost of $14$g instead of $22$g.

\subsection{Task: Dragging}

In this task, a box object of uniform density needs to be dragged on a planar surface from an initial 3D pose (2D translation and 1D rotation $[x_0, y_0, \phi_0]$) to a target goal pose $[x_g, y_g, \phi_g]$ that is sampled from a distribution every time the task is executed.
Dragging means the robot end-effector pushes the object against the workspace and drags the object along a 2D plane while maintaining contact with the object.
There are $3$ parameters varied and estimated in this task: the torsional friction of the robot-object contact, torsional friction of the object-table contact, and the mass of the object.

\begin{figure}[!tb]
    \centering
    \includegraphics[width=0.7\linewidth]{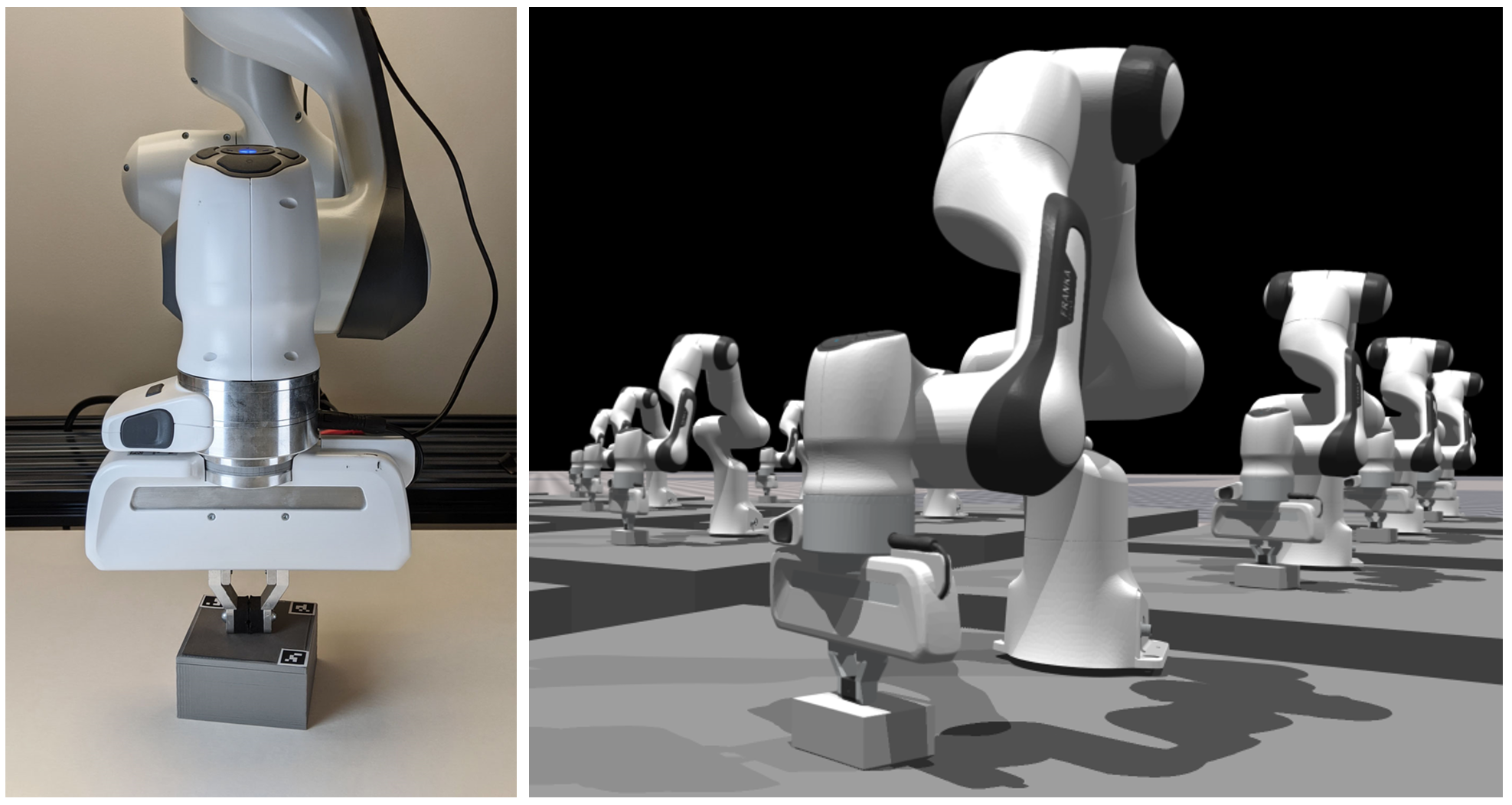}
    \caption{
        Box dragging task in real world (left) and simulation (right).
        In this task, the robot must drag the box to different goal poses on a flat surface.
        Variations across the surface material, box material, and box mass lead to different slippage behaviors.
        }
    \label{fig:dragging}
\end{figure}

The task cost is the weighed sum of the magnitude of the translation difference and the absolute value of the angular difference between the final and the goal object pose: 
$J = [\|[\hat{x}_g - x_g, \hat{y}_g - y_g]\|_2, |\hat{\phi} - \phi|]^\top w$.
The weights are chosen such that each term is roughly normalized to a ratio of $3:1$ for rotational vs. translational error across the optimization.
See Figure~\ref{fig:dragging} for an illustration of the task in both simulation and the real world.

The parameter space for the task policy consists of two 3D waypoints in the frame of the object.
The first waypoint indicates where the robot gripper makes contact with the top surface of object, and the second waypoint is where the robot gripper moves to.
The trajectory in between the end waypoints is generated via min-jerk interpolation, and the robot gripper is controlled via Cartesian end-effector impedance control.

The parameter space for the exploration policy similarly consists of two waypoints, but they are more constrained than the task waypoints, so the exploration trajectories are shorter than the task trajectories.

Instead of using analytical dynamics models as in the LQR and pouring tasks, here we use a physics simulator.
Specifically, we Isaac Gym~\footnote{\url{https://developer.nvidia.com/isaac-gym}}, a GPU-accelerated robotics simulator~\cite{liang2018gpu}, which allows us to run many simulations in parallel on a single GPU.
We simulate $20$ robots in parallel at a time step of $\Delta t=0.01$s, and this achieves roughly $100$ FPS on a single Nvidia GTX 1080 Ti GPU.

\subsubsection{Optimizers}
For both TrajOpt and SimOpt we use the episodic variant of Relative Entropy Policy Search (REPS)~\cite{peters2010relative}.
This is a derivative-free optimization algorithm that maintains the current optimal parameters as a multivariate normal distribution, and it updates the mean and covariance of the distribution at every optimization step subject to a KL-divergence constraint.

\textbf{Trajectory Optimization.}
The initial mean trajectory REPS uses consists of the first waypoint right above the center of the object, and a second waypoint to coincide with the goal delta pose for the object.
This initial trajectory doesn't work in most of the cases, because, depending on the friction and mass values, the object will slip and rotate different amounts, so its contact with the robot end-effector is not rigid.
REPS for TrajOpt converges within $10$ iterations.

\textbf{Simulation Optimization.}
Given a trajectory of object poses and robot actions during exploration, we use REPS to find the $\hat{\theta}$ in simulation that generates the trajectory closest to the observations.
The initial mean of the dynamics model parameters are sampled from the wide distribution $\Theta$, while the initial covariance is set wide enough to sufficiently cover $\Theta$.
At every REPS iteration, sampled $\theta$'s are used to form the dynamics model $f_\theta$, which is then used to playback the recorded exploration trajectory.
The translation and rotational differences between each simulation's object poses and the observed object poses are used to form a weighted sum cost similar to the one in trajectory optimization.
REPS for SimOpt also converges within $10$ iterations.

\textbf{Training the Exploration Policy.}
For optimizing the exploration policy, we first experimented with REPS, but taking the full expectation of $\Psi$ is too slow in practice.
Instead, like with the pouring task, we use finite difference in simulation to directly perform gradient updates and estimate the gradients by taking a small batch of samples.
To make multi-dimensional finite difference more accurate and efficient, we sample small perturbations around the input variable, evaluate the function at those perturbations, and fit a plane to estimate the gradient.

Since we want the task to generalize across a distribution of task parameters, evaluating $J$ in Algorithms~\ref{alg:deploy} and~\ref{alg:eval} requires estimating an expectation as well.
To reduce the nested sampling of $J$ and $\Theta$ during finite difference, we split the gradient $\E_{\theta} \nabla_{\pi_e}\Psi(\pi_\tau(\hat{\theta}), \theta)$ into two parts via chain rule and estimate them separately.
The first term is the gradient of the task cost w.r.t. the simulation parameters evaluated around the estimated simulation parameters: $\nabla_{\hat{\theta}}\E_{\theta}\Psi(\pi_\tau(\hat{\theta}), \theta)$.
The second is the gradient of the estimated simulation parameters w.r.t to the exploration policy: $\nabla_{\pi_e}\E_{\theta} \hat{\theta}$.
Gradient updates were performed via the Adam optimizer.

Figure~\ref{fig:expopt} shows the task regret ratio of task-agnostic vs. task-oriented exploration policy during training.
Similar to previous cases, the task-oriented exploration policy achieves lower task regret than the task-agnostic exploration policy.

\subsubsection{Real-world Evaluations}

\begin{figure}[!tb]
    \centering
    \includegraphics[width=0.9\linewidth]{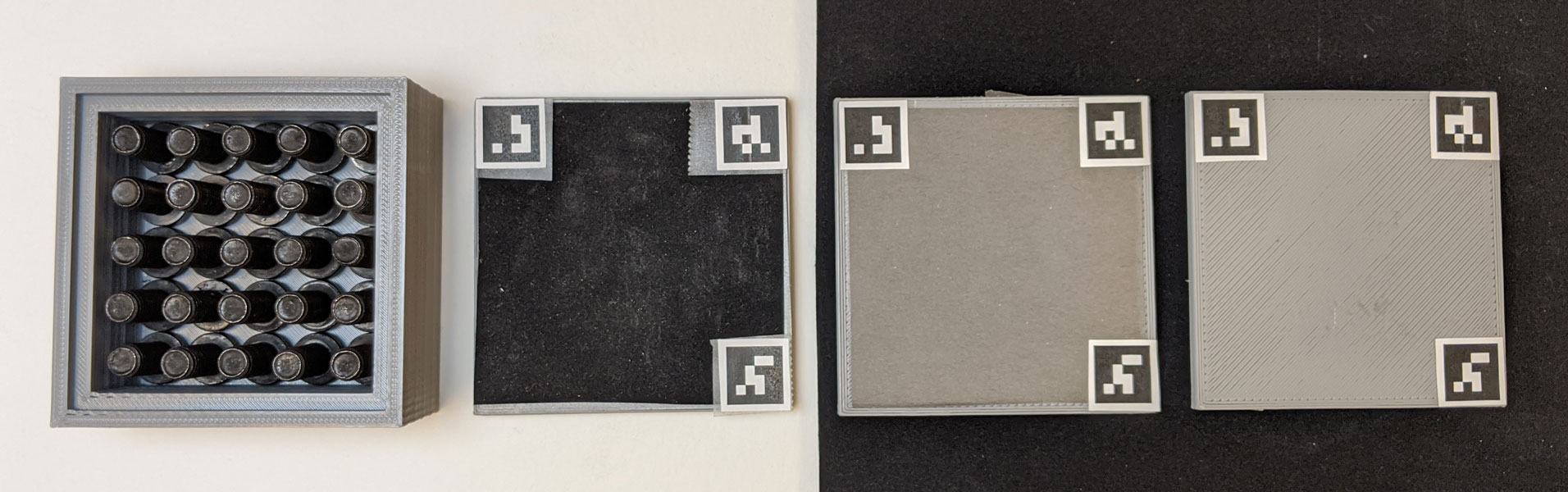}
    \caption{Real-world box dragging setup. The 3D printed box has a cavity with a removable lid, so its mass can be changed. We tested $3$ box top materials (PLA plastic, construction paper, and felt), $2$ table surface materials (construction paper and felt), and $2$ different box masses.}
    \label{fig:box_materials}
\end{figure}

To evaluate the two exploration policies in the real-world, we 3D printed a box with the exact dimensions as the one used in simulation.
The box has a cavity with a removable lid, so we can vary the box's mass.
To vary the friction parameters of the robot-object and object-table contacts, we attached different sheet materials to both the top of the box and the top of the table surface.
We also attached AprilTags~\cite{wang2016iros} to the top of the lid to estimate the initial and final poses of the box.
See Figure~\ref{fig:box_materials} for the box and the different materials used. 
In total, we experimented with $2$ sheet materials for the table surface, $3$ materials for the top of the box, and $2$ different masses for the box.

\begin{figure}[!tb]
    \centering
    \includegraphics[width=0.65\linewidth]{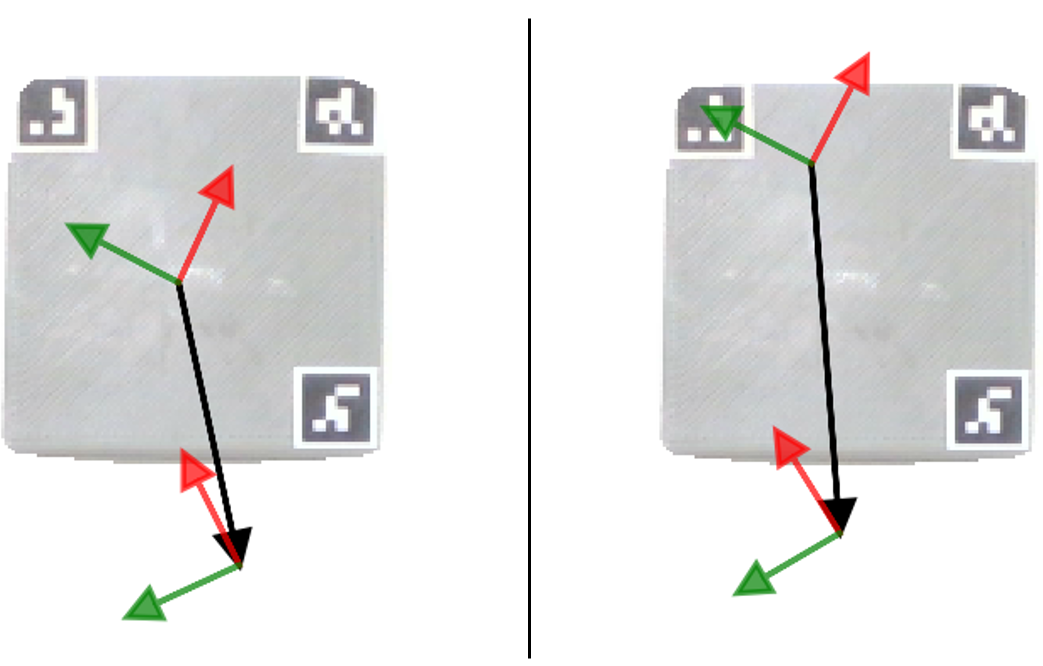}
    \caption{
        Learned task-agnostic exploration trajectory (left) and task-oriented exploration trajectory (right) for the dragging task.
        The dragging trajectories consist of start and end 3D way points for the robot's end-effector, and they're denote by the red and green axes.
        While both trajectories have comparable translation and rotation magnitudes, the task-oriented exploration trajectory begins further away from the object's center of mass.
        This leads to an object trajectory that is more sensitive to the torsional friction between the object and the table surface.
    }
    \vspace{3pt}
    \label{fig:box_explore_trjas}
\end{figure}

\begin{figure}[!tb]
    \centering
    \includegraphics[width=0.49\linewidth]{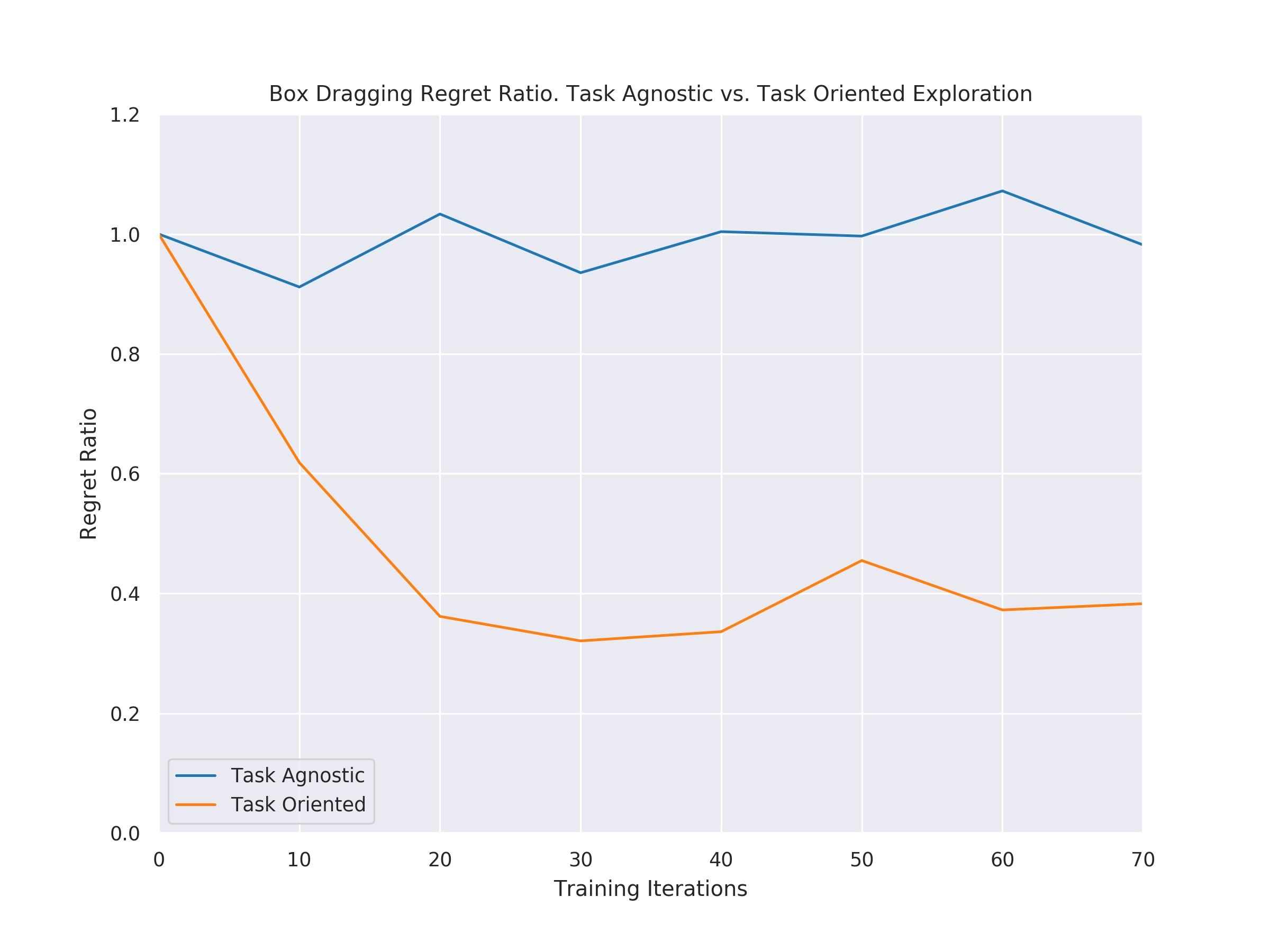}
    \includegraphics[width=0.49\linewidth]{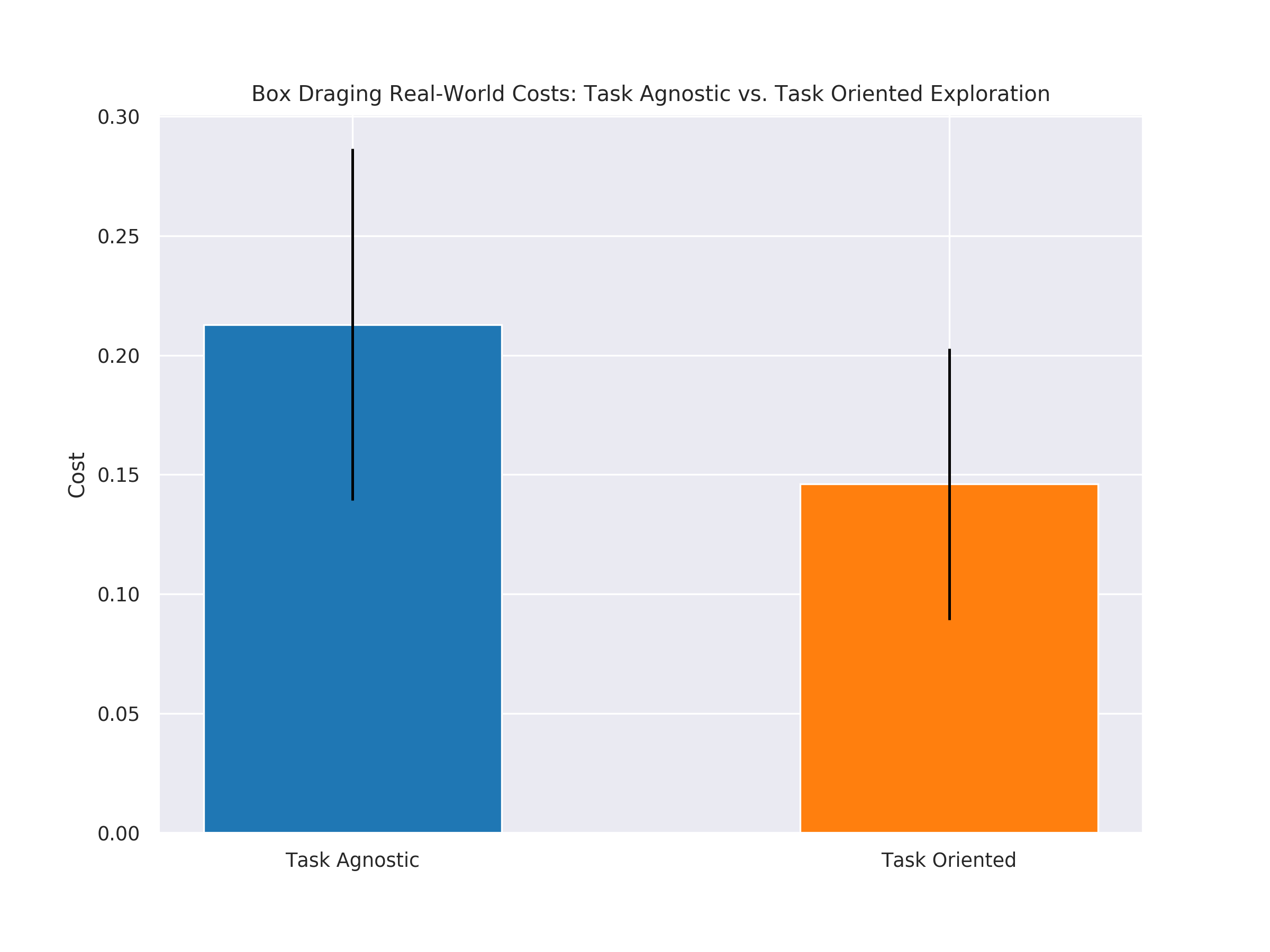}
    \vspace{-25pt}
    \caption{
        Box dragging task results. 
        Left: Regret ratio of task-agnostic vs. task-oriented exploration policy during training. 
        Right: Real-world task execution costs using learned task-agnostic vs. task-oriented exploration policies. Costs are aggregated over $48$ trials per method with different box masses and surface materials.
        }
    \vspace{-3pt}
    \label{fig:expopt}
\end{figure}

We evaluate each set of parameters with $3$ different goal poses with $2$ trials each, and we removed all trials during which the robot was not able to move the box at all.
This happens when the friction of the object-table contact is much greater than that of the robot-object contact.
In total, there are $48$ trials used for evaluating each of the task-oriented and task-agnostic exploration policies.
See Figure~\ref{fig:box_explore_trjas} for a visualization of the learned exploration trajectories and Figure~\ref{fig:expopt} for task performance during training and real-world evaluations.
The Task-Oriented exploration policy led to smaller mean and standard deviation of costs than the Task-Agnostic policy, which did not improve over random exploration.
For this task, random exploration already leads to reasonable Sys-ID, and task-oriented information is needed for further improvements.

\subsection{Discussions}
In both the pouring and box dragging task, and during both training the exploration policy and testing it in the real world, the parameters identified by task-oriented exploration led to better final task performance than ones identified by task-agnostic exploration.
This behavior is observed with tasks using both analytical models (LQR, pouring tasks) and black-box models (box dragging), and with discrete exploration actions (pouring tasks) and continuous ones (LQR, box dragging).

The advantage of task-oriented over task-agnostic exploration is due to the limited exploration budget.
Exploration policies have a finite time horizon, and during training regularization terms are added to prevent the policy from incurring high-cost states or actions (e.g. actions or states that are too large).
As a result, there is a need to explore more about system parameters to which the task cost is more sensitive.
This is reflected by weighting the dynamics prediction error by the task cost in LQR (Equation~\ref{eq:lqr_obj}), measuring the mass of the task-relevant cup more in pouring (Figure~\ref{fig:pouring_training}), and exploring from an initial contact further away from the object's center of mass in box dragging (Figure~\ref{fig:box_explore_trjas}).

Because the performance of an exploration policy during training is evaluated as an expectation over a wide distribution of system parameters as well as task goals, the learned exploration policies can be applied to different systems and generalize across the task distribution.
This benefit also points to a limitation of our approach, which is that optimizing the exploration policies requires $3$ layers of nested sampling.
They include taking samples for finite difference approximation, system parameters, and task goals.
As a result, the gradient $\nabla_{\pi_e} \E_\theta \Psi$ can be slow to evaluate when full dynamics simulation is used.
However, tasks with many model parameters typically only have a small subset of parameters that significantly affect task performance.
As such, the benefits of the task-oriented approach will be more apparent in high-dimensional tasks, where the effective task-oriented dimenaionalty is much lower than the task-agnostic dimensionality.

%% file: includes/6_conclusion.tex
\section{Conclusion}
In this paper, we proposed, analyzed, and implemented a framework of learning active task-oriented exploration policies to improve task performance in the real world and bridge the sim-to-real gap.
The learned exploration policy works across system parameters and task goals, so it can be applied to different variations of the task without retraining.
We instantiated the framework with three experiments using analytical and full dynamics simulation models.
Across all experiments we observed that task-oriented exploration leads to better task performance than task-agnostic exploration.


%% file: includes/7_ack.tex
\section*{Acknowledgements}
\scriptsize{
The authors thank Shivam Vats, Steven Lee, and Kevin Zhang for their suppport in this project.
This work is funded by the NSF Graduate Research Fellowship Program Grant No. DGE 1745016, NSF Award No. CMMI-1925130, the Office of Naval Research Grant No. N00014-18-1-2775, and ARL grant W911NF-18-2-0218 as part of the A2I2 program.
}

\normalsize{}

%% file: includes/8_appendix.tex
\begin{appendices}

\section{LQR Experiment}

\subsection{System Parameters}
\begin{itemize}
    \item $n = 6$, $m = 3$
    \item $\theta \sim \mathcal{N}([0.9, 0.9, 0.9, 0.6, 0.6, 0.6], 0.2I)$. Samples are clipped at a magnitude of $1.1$.
    \item Observation noise distribution: $\mathcal{N}(0, 0.05)$
    \item Dynamics noise distribution: $\mathcal{N}(0, 0.05)$
\end{itemize}

\subsection{Task Parameters}
\begin{itemize}
    \item $Q = diag([100, 100, 10, 10, 10, 1])$
    \item $R = diag([0.1, 0.1, 0.1])$
\end{itemize}

\subsection{Adam Optimizer Hyperparameters}
\begin{itemize}
    \item $\alpha = 10^{-4}$
    \item $\beta_1 = 0.9$
    \item $\beta_2 = 0.999$
    \item $\epsilon = 10^{-8}$
    \item weight decay $= 0.1$
    \item batch size $= 70$
\end{itemize}

\section{Pouring Experiment}

\subsection{Parameters}

Initial water masses for both cups are drawn from a uniform distribution in the range of $[150, 300]$g.

Noise sampled from $\mathcal{N}(0, 30)$ (unit in g) and clipped at $\sigma$ was added to each mass measurement.

Noise sampled from $\mathcal{N}(0, 5)$ (unit in g) and clipped at $\sigma$ was added to each pouring outcome simulation.

The range used for finite difference perturbations is $0.05$.

\subsection{Real-world Mass Measurement Errors}

\begin{figure}[!b]
    \centering
    \includegraphics[width=\linewidth]{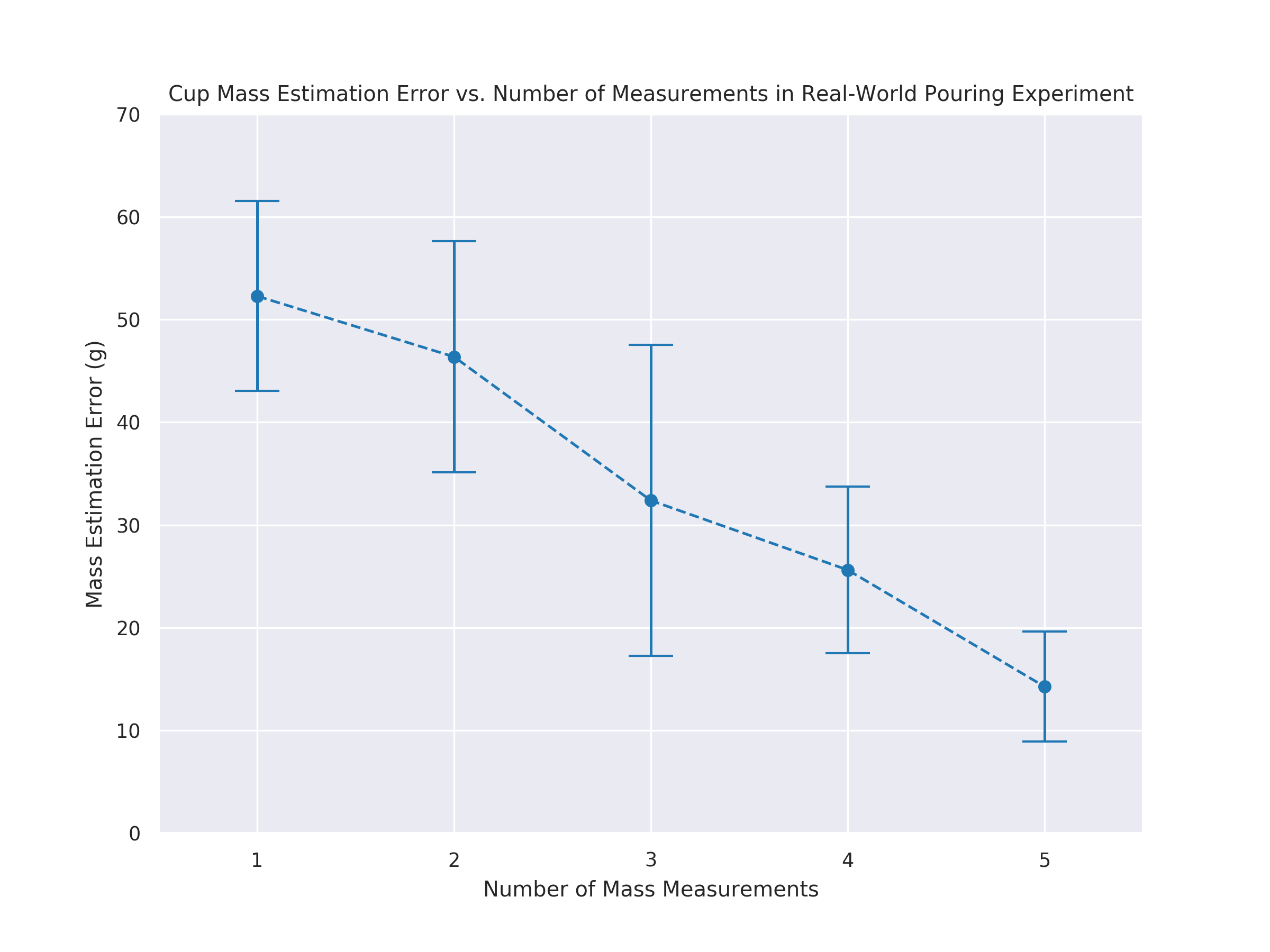}
    \vspace{-20pt}
    \caption{
        Cup mass estimation error vs. number of mass measurements.
        Data aggregated over all pouring experiment trials.
        Length of error bars denote one standard deviation.
    }
    \label{fig:mass_measurement_err}
\end{figure}

Figure~\ref{fig:mass_measurement_err} plots how the cup mass estimation errors decrease as the number of measurements increase.
With $5$ measurements, the mean estimation error decreases from the initial $50$g to about $15$g.

\subsection{Adam Optimizer Hyperparameters}
\begin{itemize}
    \item $\alpha = 5\cdot 10^{-3}$
    \item $\beta_1 = 0.9$
    \item $\beta_2 = 0.999$
    \item $\epsilon = 10^{-8}$
    \item weight decay $= 0$
    \item batch size $= 100$
\end{itemize}

\section{Box Dragging Experiment}

\subsection{Box Dragging Task Visualization}
\begin{figure}[!h]
    \centering
    \includegraphics[width=0.6\linewidth]{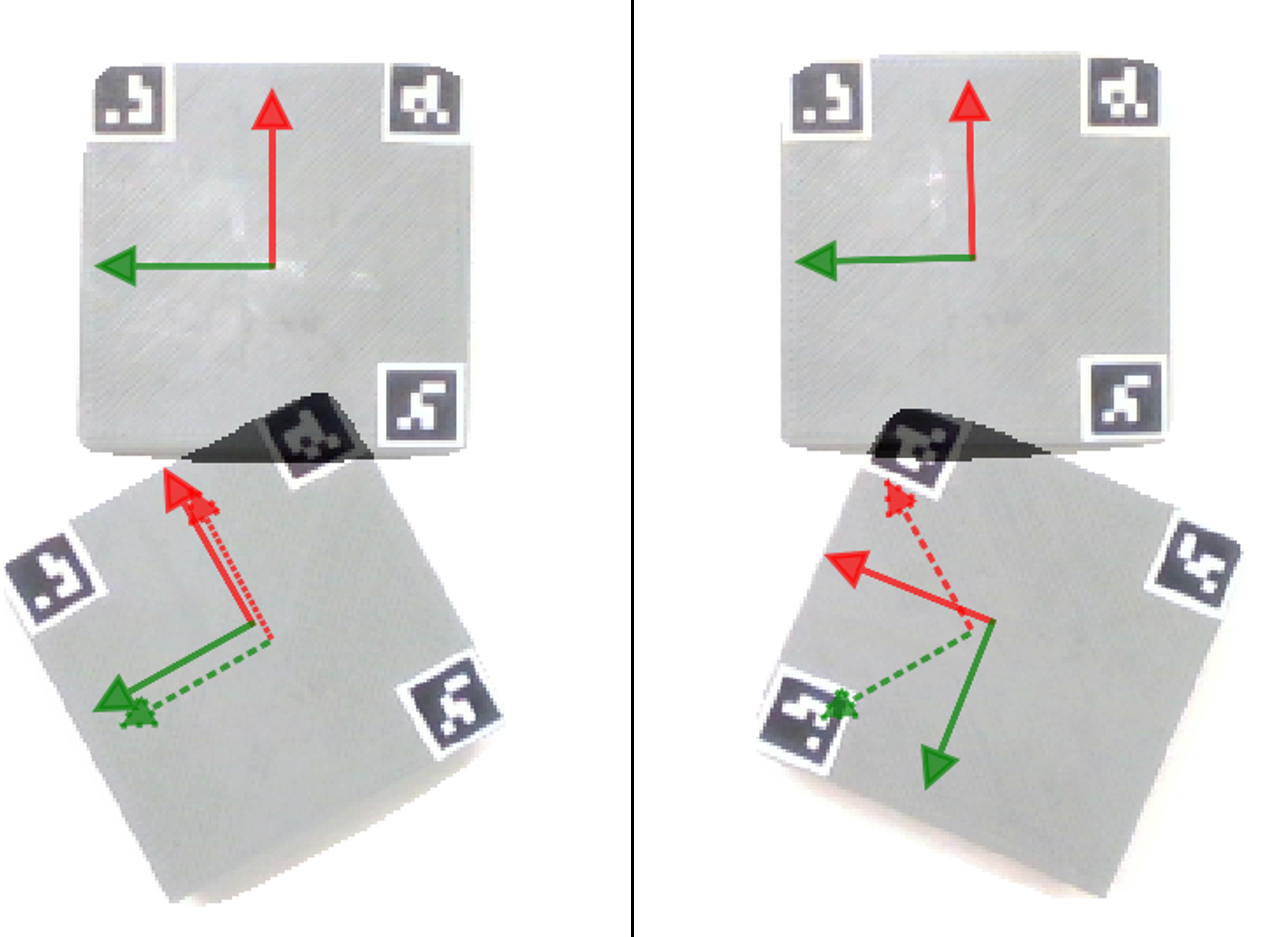}
    \caption{
        Two example dragging trajectories for the same goal pose. 
        In both trials the boxes start at the top. 
        The solid arrows indicate the initial and final box poses, and the dashed arrows indicate the goal pose.
        On the left the goal pose and the final pose almost align, while on the right there is a big difference in the final and goal pose angles.
    }
    \vspace{-6pt}
    \label{fig:drag_box}
\end{figure}

\subsection{Dragging Task Goal Distribution}
\begin{itemize}
    \item Translation: $[x_g, y_g] \sim \mathcal{N}(0, 0.3)$.
    \item Rotation: $\theta_g \sim \mathcal{N}(0, 50^\circ)$
\end{itemize}

Samples are clipped at $1.5\sigma$.

\subsection{System Parameters}
The range boundaries form the support of the uniform distribution $\Theta$.
Friction refers to torsional friction.
The priors are the initial distributions used by REPS during SimOpt.
During REPS, the parameter samples are clipped at $2\sigma$ and by the parameter's corresponding range.
Delta is the range used for finite difference perturbations.

\begin{table}[!h]
\begin{tabular}{l|lll}
                                & Range                       & Prior                                & Delta     \\ \hline
Robot-Object Friction & $[0.01, 0.4]$               & $\mathcal{N}(0.15, 0.2)$             & $0.01$    \\
Object-Table Friction & $[10^{-3}, 4\cdot 10^{-3}]$ & $\mathcal{N}(2\cdot 10^{-3}, 0.06)$ & $10^{-4}$ \\
Object Mass                     & $[0.05, 0.5]$               & $\mathcal{N}(0.15, 0.3)$             & $0.01$   
\end{tabular}
\end{table}

\subsection{Dragging Trajectory Optimization}
A dragging trajectory consists of two way points that are interpolated via min-jerk interpolation.
The waypoints are specified in the object frame, with the origin point coinciding with the object center.
The initial waypoints used for both the task policy and the exploration have start and end poses as follows:

\begin{itemize}
    \item $[x_0, y_0]\sim\mathcal{N}(0, 0.1)$, $\phi_0\sim\mathcal{N}(0, 20^\circ)$
    \item $[x_{T_\tau}, y_{T_\tau}, \phi_{T_\tau}]\sim [0.1, 0, 0]$
\end{itemize}

Samples of the first waypoints are clipped at $\sigma$ and also by the boundaries of the box, so the robot is guaranteed to make contact with the box initially.

The range used for finite difference perturbations are: $[\Delta x, \Delta y, \Delta z] = [2\cdot 10^{-3}, 2\cdot 10^{-3}, 0.1]$

\subsection{REPS Iteration Counts}
\begin{itemize}
    \item SimOpt: $8$
    \item TrajOpt: $5$
\end{itemize}

\subsection{Adam Optimizer Hyperparameters}
\begin{itemize}
    \item $\alpha = 5\cdot 10^{-3}$
    \item $\beta_1 = 0.9$
    \item $\beta_2 = 0.999$
    \item $\epsilon = 10^{-8}$
    \item weight decay $= 0.01$
    \item batch size $= 5$
\end{itemize}

\section{Relative Entropy Policy Search}

The episodic variant of REPS works as follows:

Let $z$ be the optimization variable and $R(z)$ be the reward function (for minimizing costs, set rewards to the negative costs).
REPS is initialized with a normal distribution over the optimization variable $\mathcal{N}(\mu_z, \Sigma_z)$.
At every REPS iteration, we draw $N$ samples from the current normal distribution over $z$ which gives us a set of $z_n$'s.
Then, each sample is evaluated for a reward $R_n$, which are used to update the distribution over $z$.
The update is performed by first computing the temperature parameter $\eta$ from the KL-divergence constraint $\epsilon$ by minimizing the following objective:
\begin{equation}
    \eta^* = \argmin_\eta \eta\epsilon + \eta\log\frac{1}{N}\sum_{n=1}^N e^{R_n/\eta}
\end{equation}

The new mean and covariance are:
\begin{align}
    d_n &= e^{\frac{R_n}{\eta}}\\
    \mu_z &= \frac{\sum_{n=1}^N d_n z_n}{\sum_{n=1}^N d_n} \\
    \Sigma_z &= \frac{\sum_{n=1}^N d_n (z_n - \mu)(z_n - \mu)^\top}{\sum_{n=1}^N d_n}
\end{align}

We use $\epsilon=1$ for all experiments.

\section{Finite Difference via Plane Fitting}

The finite difference approximation used during training the pouring and box dragging exploration policies is detailed in Algorithm~\ref{alg:fd}.
The inputs to the algorithm are:
\begin{itemize}
    \item $f$ - the function to be differentiated.
    \item $x$ - the input around which $f$ is differentiated.
    \item $\Delta$ - the variance around which to sample perturbations.
    \item $l$ and $u$ - the lower and upper bounds for $x$. This is useful for bounding the input perturbations, for example when $x$ is a probability between $0$ and $1$.
    \item $N$ - the number of samples to use.
\end{itemize}

\floatstyle{spaceruled}
\restylefloat{algorithm}
\begin{algorithm}[!h]
\caption{Finite Difference via Plane Fitting}
\label{alg:fd}
\begin{algorithmic}[1]
    \renewcommand{\algorithmicrequire}{\textbf{Input:}}
    \renewcommand{\algorithmicensure}{\textbf{Output:}}
    \REQUIRE $f, x, \Delta, l, u, N$
    \STATE $X \leftarrow$ $N$ samples from $x + \max(\min(\mathcal{N}(0, \Delta), u), l)$
    \STATE $F \leftarrow [f(X_1), \hdots, f(X_N)]$ 
    \STATE $\Bar{X} \leftarrow \frac{1}{N}\sum_{n=1}^N X_n$
    \STATE $\Bar{F} \leftarrow \frac{1}{N}\sum_{n=1}^N F_n$
    \STATE $\nabla_x f(x) \leftarrow (X - \Bar{X})^\dagger (F - \Bar{F})$
    \RETURN $\nabla_x f(x)$
\end{algorithmic}
\end{algorithm}

The $\dagger$ symbol denotes taking the pseudo-inverse.

\end{appendices}